\documentclass[10pt,twocolumn,letterpaper]{article}

\usepackage{wacv}
\usepackage{times}
\usepackage{epsfig}
\usepackage{graphicx}
\usepackage{amsmath}
\usepackage{amssymb}
\usepackage{multirow, makecell}
\usepackage{array}
\usepackage[numbers,sort]{natbib}
\usepackage{lineno}
\usepackage{enumerate}
\usepackage{amsfonts}
\usepackage{gensymb}
\usepackage{bm}
\usepackage{diagbox}
\usepackage{tabularx}
\usepackage{rotating}
\usepackage{xmpmulti}
\usepackage{arydshln}
\usepackage[dvipsnames]{xcolor}
\usepackage{stmaryrd}
\usepackage{relsize}
\usepackage[accsupp]{axessibility}

\def\wacvPaperID{50} 

\wacvalgorithmstrack   

\wacvfinalcopy 

\ifwacvfinal
\usepackage[breaklinks=true,bookmarks=false]{hyperref}
\else
\usepackage[pagebackref=true,breaklinks=true,colorlinks,bookmarks=false]{hyperref}
\fi

\usepackage{cleveref}

\begin{document}

\title{Modality Mixer for Multi-modal Action Recognition}

\author{Sumin Lee \quad Sangmin Woo \quad Yeonju Park \quad Muhammad Adi Nugroho \quad Changick Kim\\
KAIST\\
{\tt \small \{suminlee94, smwoo95, yeonju29, madin, changick\}@kaist.ac.kr} 
}


\maketitle

\begin{abstract}
    In multi-modal action recognition, it is important to consider not only the complementary nature of different modalities but also global action content.
    In this paper, we propose a novel network, named Modality Mixer (M-Mixer) network, to leverage complementary information across modalities and temporal context of an action for multi-modal action recognition.
    We also introduce a simple yet effective recurrent unit, called Multi-modal Contextualization Unit (MCU), which is a core component of M-Mixer.
    Our MCU temporally encodes a sequence of one modality (\eg, RGB) with action content features of other modalities (\eg, depth, IR).
    This process encourages M-Mixer to exploit global action content and also to supplement complementary information of other modalities.
    As a result, our proposed method outperforms state-of-the-art methods on NTU RGB+D 60, NTU RGB+D 120, and NW-UCLA datasets. 
    Moreover, we demonstrate the effectiveness of M-Mixer by conducting comprehensive ablation studies.
\end{abstract}

\section{Introduction} 
\label{chap:intro}
Humans experience their surroundings through a combination of various modality data, such as audio, sight, and touch.
According to the recent advancements in sensor technology, multi-modal learning has attracted much research interest in the field of computer vision. 
Toward this direction, for video action recognition, many multi-modal methods have been developed, which achieved higher performance than other methods based on a single modality.

\begin{figure}[t]
    \centering
    \includegraphics[width=0.5\textwidth]{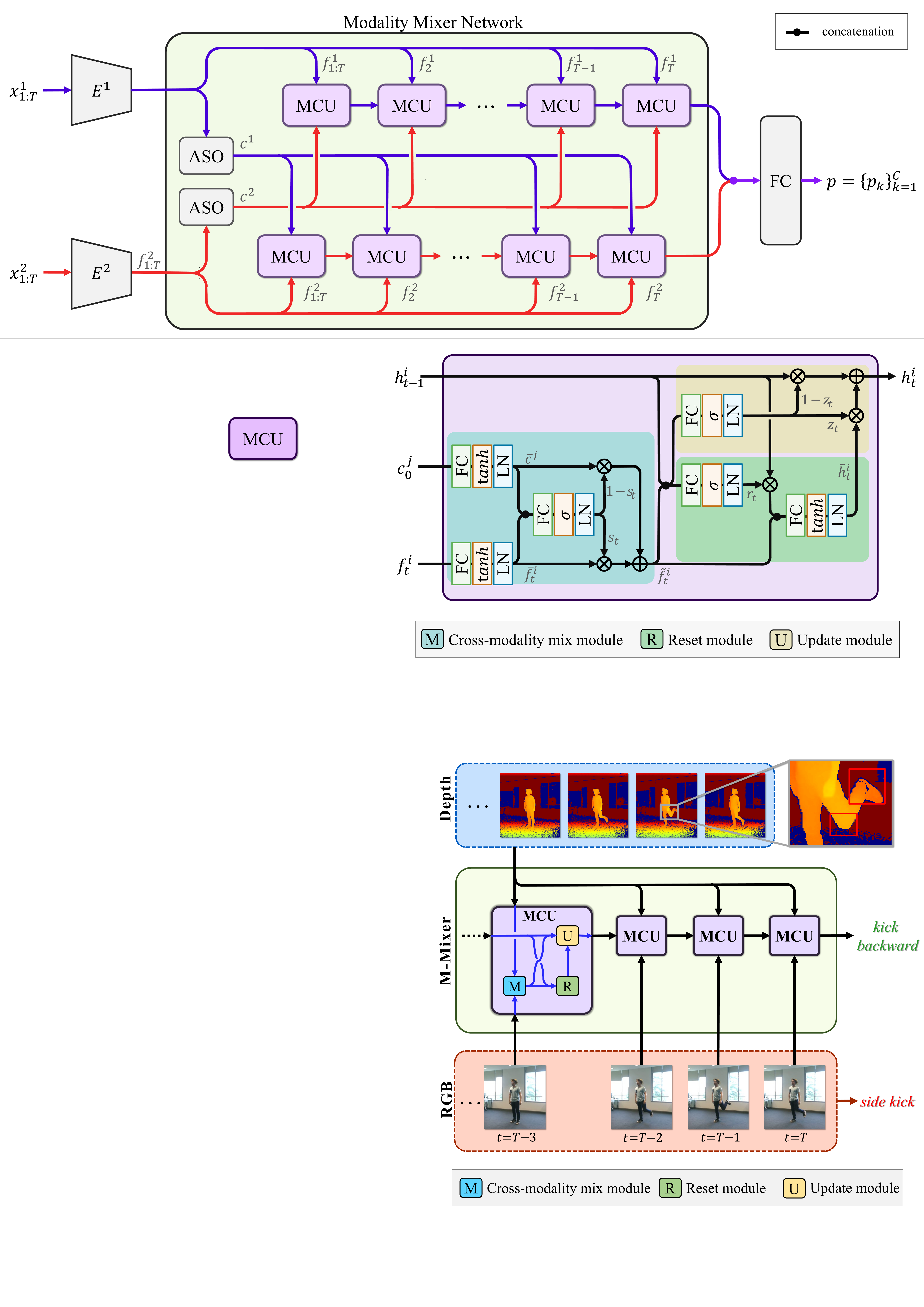}
    \caption{\textbf{Multi-modal Action Recognition with Modality Mixer (M-Mixer) network.}
    Solely relying on appearance data of the RGB modality, the action `kick backward' is easily misclassified to `\textcolor{red}{side kick}'. 
    On the other hand, our MCU supplements the information on foot orientation from the depth image: the foot is going behind the knee (the more yellowish the color is, the closer it is to the camera).
    As such, M-Mixer correctly classifies the action as `\textcolor{Green}{kick backward}.'
    Overall, MCU performs temporal encoding of the RGB sequence, while augmenting it with complementary information of depth content feature.
    Here, we assume to use RGB and depth input and depict only RGB stream for the sake of clarity.
    }
    \label{fig:intro}
\end{figure}

Earlier studies on action recognition mostly relied on a single RGB modality~\cite{3d_temp:i3d, 3d_temp:r3d, 3d_temp:nonlocal, 3d_temp:slowfast}, which are focused on spatio-temporal modeling.
Lately, many models based on multi-modality have been developed to integrate information of different modalities such as RGB, optical flow, and depth~\cite{rgbd_3, mstd, rgbd_4, shahroudy2017deep, liu2018viewpoint, garcia_admd, garcia_eccv, garcia_dmcl, mmar_lstm}.
Due to the different properties of sensors, each modality possesses different key characteristics that contribute to the overall action recognition.
Specifically, while RGB images provide visual appearances, depth data contains the 3D structure of 2D frames, which is complementary to RGB modality.
For example, as illustrated in Fig.~\ref{fig:intro}, the action `kick backward' may be incorrectly predicted as `side kick' when employing only the RGB modality~\cite{ntu120}.
However, the depth data can indicate that the foot is going behind the knee, leading to the correct action class `kick backward'.
Therefore, in order to distinguish a correct action class, it is necessary to integrate complementary information from multi-modal data as well as temporal encoding of given videos.

In this paper, we propose a novel network, Modality Mixer (M-Mixer), which leverages two important factors for multi-modal action recognition: 1) complementary information across modalities and 2) temporal context of action.
Taking feature sequences of multiple modalities as inputs, our M-Mixer temporally encodes each feature sequence with action content features of other modalities.
The action content features include modality-specific information and the overall activity of videos.
We also introduce a simple yet effective recurrent unit, called Multi-modal Contextualization Unit (MCU), which adaptively integrates a modality sequence and action content features.
Our M-Mixer network employs a distinct MCU for each modality.
As each MCU is dedicated to a specific modality, we describe our MCU in detail from an RGB perspective, as illustrated in Fig.~\ref{fig:intro}.
Our MCU consists of three modules: cross-modality mix module, reset module, and update module.
Concretely, given an RGB feature at certain timestep and context features of other modalities, our cross-modality mix module models their relationship and adaptively integrates them by weighted summation.
By doing so, MCU enables the network to exploit complementary information across modalities and global action content during temporal encoding.
Then, reset and update modules learn the relationships between the integrated feature of the current timestep and thehidden state of the previous timestep.
Based on MCU, our M-Mixer network assimilates more richer and discriminative information from multi-modal sequences for action recognition.
Note that, our M-Mixer is not limited to only two modalities and is applicable to more modalities.

We perform extensive experiments on three benchmark datasets (\ie, NTU RGB+D 60~\cite{ntu60}, NTU RGB+D 120~\cite{ntu120}, and Northwestern-UCLA (NW-UCLA)~\cite{nwucla}).
Our M-Mixer network achieves the state-of-the-art performance of 90.77\%, 90.12\%, and 94.43\% on NTU RGB+D 60, NTU RGB+D 120, and NW-UCLA datasets, respectively.
Also, we empirically show that our M-Mixer model can be extended to more than two modalities.
Through extensive ablation experiments, we demonstrate the effectiveness of the proposed M-Mixer network.

Our main contributions are summarized as follows:
\renewcommand\labelitemi{\tiny$\bullet$}
\begin{itemize}
\item We investigate how to take two important factors into account for multi-modal action recognition: 1) complementary information across modality, and 2) temporal context of an action.
\item We introduce a novel network, named M-Mixer, with a new recurrent unit, called MCU. By effectively modeling the relation between a sequence of one modality and action contents of other modalities, our MCU facilitates M-Mixer to exploit rich and discriminative features.
\item We evaluate the performance of multi-modal action recognition on three benchmark datasets. Experimental results show that M-Mixer outperforms the state-of-the-art methods. Moreover, we demonstrate the effectiveness of the proposed method by conducting comprehensive ablation studies.
\end{itemize}


\section{Related Work}
\label{chap:rw}
Video action recognition is one of the most representative tasks in the field of video understanding.
Thanks to the emergence of deep learning, video action recognition has made significant progress over the last decade.
Early deep learning models~\cite{2stream:2stream, 2stream:fusion, 2stream:tsn} were developed with a two-stream structure in which each stream captures the appearance and motion data of videos, respectively.
Due to the high cost of computing accurate optical flow, other works~\cite{rgb3d:d3d, rgb3d:dance, rgb3d:fixed, rgb3d:flow, rgb3d:guided, rgb3d:mars} studied to learn to mimic motion features from only RGB sequences.
Stroud \etal~\cite{rgb3d:d3d} and Crasto \etal~\cite{rgb3d:mars} suggested learning algorithms that distill knowledge from the temporal stream to the spatial stream in order to reduce the two-stream architecture into a single-stream model.
Other studies ~\cite{rgb3d:fixed,rgb3d:dance,rgb3d:flow} introduced modules to explore motion information in a unified network.
After then, 3D convolution networks~\cite{3d_temp:i3d, 3d_temp:r3d, 3d_temp:nonlocal, 3d_temp:slowfast} were proposed for action recognition, which led to significant performance improvements.
Among them, SlowFast network~\cite{3d_temp:slowfast} consists of two pathways for dealing with two different frame rates to capture spatial semantics and motion.

\begin{figure*}[t]
    \centering
    \includegraphics[width=\textwidth]{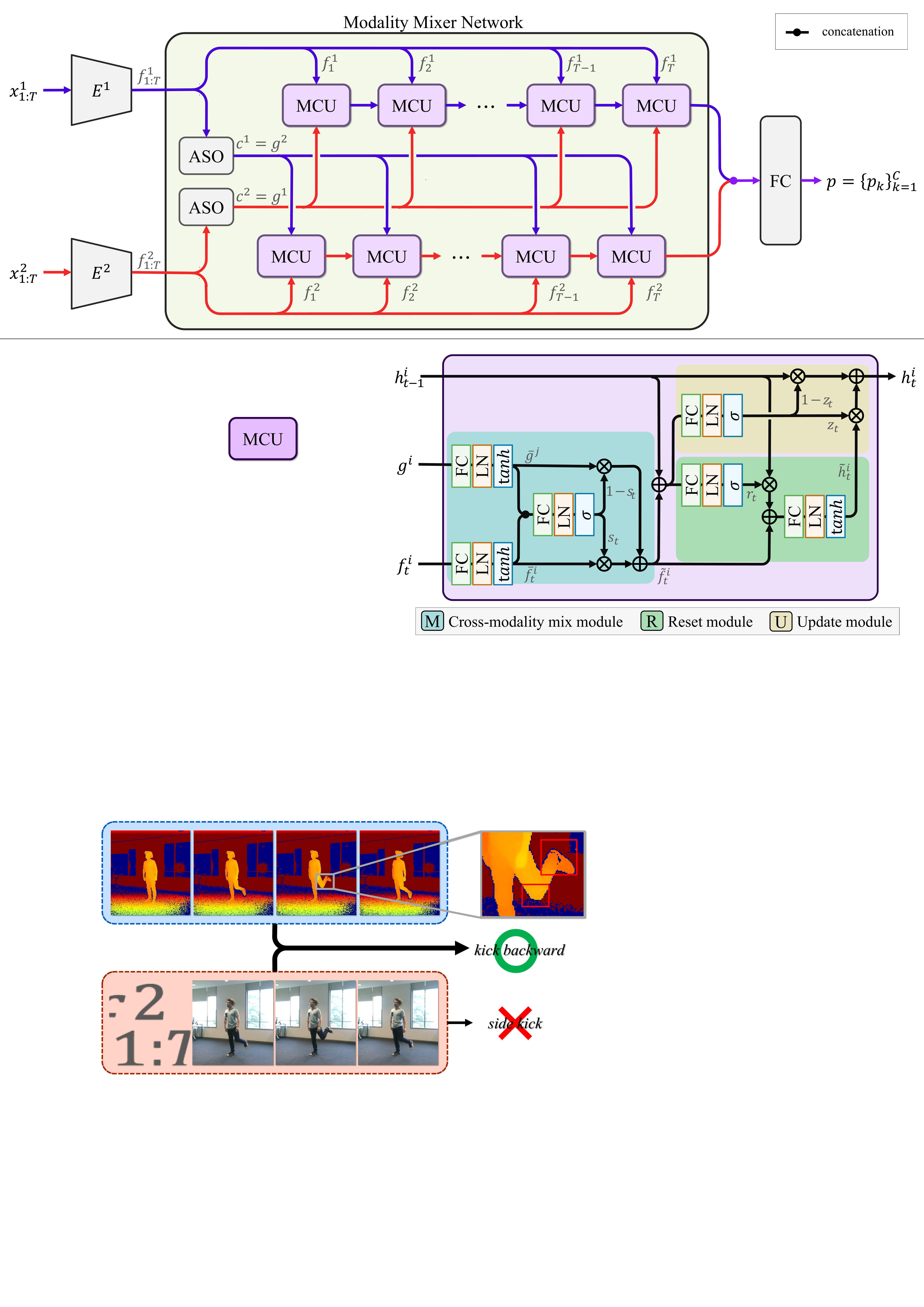}
    \caption{\textbf{Modality Mixer network.} 
    We illustrate an example of using two modalities in this figure. M-Mixer network consists of Multi-modal Contextualization Unit (MCU) and Action Summarizing Operator (ASO).
    Our M-Mixer takes a feature sequence $f^{i}_{1:T}$, derived from a feature extractor $E^i$ by a frame sequence $x^i_{1:T}$, as an input.
    First, ASO computes an action content feature $c^i$ from $f^i_{1:T}$.
    Then, our MCU temporally encodes $f^{i}_{1:T}$ with a cross-modality action content feature $g^{i}$, which is $c^{j}$ in this case, where $j\neq i$.
    By comparing $f^{i}_{1:T}$ with $g^{i}$ during temporal encoding, MCU considers complementary information across modalities and overall action contents of videos.
    The final probability distribution over $K$ action classes is calculated by using $h^1_T$ and $h^2_T$.
    Here, the blue and red lines indicate streams of modality 1 and 2, respectively, and the purple line represents the fusion of modalities.
    }
    \label{fig:overview}
\end{figure*}

Due to the advance of sensor technologies, action recognition in multi-modal setting has attracted research interest~\cite{skel_1, skel_2, wang2020makes, alayrac2020self, huang2021makes, caesar2020nuscenes,audio2,audio1}
RGB and depth are one of the most common combinations of modalities~\cite{rgbd_1, rgbd_2, rgbd_3, rgbd_4}.
Shahroudy \etal~\cite{shahroudy2017deep} introduced a shared-specific feature factorization network based on autoencoder structure for RGB and depth inputs.
Liu \etal~\cite{liu2018viewpoint} presented a method of learning action features that are insensitive to camera viewpoint variation.
Wang~\cite{rgbd_aaai} proposed a cooperatively trained Convolutional neural Network (c-ConvNet), which enhances the discriminative information of RGB and depth modalities.
In~\cite{mstd}, a two-stream view-invariant framework was proposed with motion stream and Spatial-Temporal Dynamic (STD) stream with RGB and depth, where late fusion technique is employed to combine outputs of these RGB and depth streams.
In~\cite{garcia_eccv, garcia_admd}, frameworks of distillation and privileged information were suggested.
Although these methods are trained with both RGB and depth data, a hallucination network of depth enables classifying actions with only RGB data.
Garcia \etal~\cite{garcia_dmcl} introduced an ensemble of three specialist networks, called the Distillation Multiple Choice Learning (DMCL) network, that works with missing modalities at inference time.
DMCL includes each specialist network for RGB, depth, and optical flow videos that collaborates and strengthens each other and employ employs the late fusion scheme.
Wang \etal~\cite{mmar_lstm} proposed a hybrid network based on CNN (\eg, ResNet50~\cite{resnet} and 3D convolution) and RNN (\eg, ConvLSTM~\cite{convlstm}) to fuse RGB, depth, and optical flow modalities.

In this paper, we explore how to fuse multi-modal data with recurrent units.
To this end, we propose a novel recurrent unit, Multi-modal Contextualization Unit (MCU), and network, Multi-modal Mixer (M-Mixer), for multi-modal action recognition.
By encoding a feature sequence with content features of other modalities, our M-Mixer network facilitates to exploring complementary information across modalities and temporal action contents.
Note that our M-Mixer network is not limited to types and the number of video modalities.



\section{Proposed Method}
\label{chap:mixer}
In this section, we first describe the overall architecture of our proposed Modality Mixer (M-Mixer) network and then explain the proposed Modality Contextualization Unit (MCU) in detail.
In Fig.~\ref{fig:overview}, the framework of our M-Mixer network is illustrated, assuming the use of two modalities.

\subsection{Modality Mixer Network}
The goal of our M-Mixer network is to generate rich and discriminative features for action recognition from videos of $N$ different modalities.
Given a video of length $T$ for the $i$-th modality, a feature extractor $E^i$ converts a sequence of frames, $x^i_{1:T}\in \mathbb{R}^{3\times T\times H\times W}$, to a sequence of features, $f^i_{1:T} \in \mathbb{R}^{d_f \times T}$, as follows:  
\begin{align}
    f^i_{1:T} &= E^i \left (x^i_{1:T} \right ),
\end{align}
where $H$ and $W$ denote the height and width of a video, and $i = 1,2,\cdots, N$.
Then, the proposed M-Mixer network takes the extracted feature sequences $f^i_{1:T}$ as inputs.

In our M-Mixer, firstly, Action Summarizing Operator (ASO) condenses an action content information $c^i \in \mathbb{R}^{d_c}$ from $f^i_{1:T}$, as follows:
\begin{align}
    c^i &= \textrm{ASO} \left (f^i_{1:T} \right ).
\end{align}
ASO can be any operation that can concentrate action content information from a feature sequence into a vector, such as max pooling, average pooling, and recurrent units.
Among several instantiations, we empirically find that average pooling performs the best for ASO.

Our M-Mixer contains $N$ MCUs that are responsible for each modality.
Each MCU encodes a feature sequence of a designated modality with content features of other modalities in a temporal manner.
In other words, an MCU for the $i$-th modality contextualizes $f^i_{1:T}$ with the $j$-th action content for all $j \in \{ 1,\cdots,N\}$, where $j \neq i$.
For these action contents, we define a cross-modality action content feature $g^i \in \mathbb{R}^{(N-1)\times d_c}$, as follows:
\begin{align}
    g^i &=  \underset{\forall j}{\mathlarger{\mathlarger{\mathlarger{\parallel}}}} c^j, \; \textrm{where} \, j \neq i.
\label{eq:gi}
\end{align}
Here, $\parallel$ indicates a vector concatenation.
If only two modalities are used, $g^i$ is equal to $c^j$.
Then, MCU takes $f^i_t$ and $g_i$ to generate hidden state $h^i_t\in \mathbb{R}^{d_h}$ as follows:
\begin{align}
    h^i_t = \,&\texttt{MCU}^i \left (f^i_t, g^i\right ),
\end{align}
where $\texttt{MCU}^i$ denotes an MCU for the $i$-th modality.
By augmenting a cross-modality action content feature $g^i$, our MCU exploits complementary information as well as global action content,

To obtain final probability distribution $p=\{p_k\}_{k=1}^K$ over $K$ action classes, we employ a fully connected layer to $h^i_T$ for $i^{\forall}$, as follows:
\begin{align}
    p = \textrm{softmax}\left (\mathbf{W}_p \,  \left ( \underset{i\forall}{\mathlarger{\mathlarger{\mathlarger{\parallel}}}} h^i_T \right ) +b_p\right ),
\end{align}
where $p_k$ is a probability of the $k$-th action class, $\mathbf{W}_p \in \mathbb{R}^{(N\times d_h) \times K}$ is a learnable matrix, and $b_p \in \mathbb{R}^{K}$ is a bias term.

To train our M-Mixer network, we define a loss function $L$ by utilizing the standard cross-entropy loss as follows:
\begin{align}
    L = \sum_{k=1}^{K} y_k\log \left (p_k \right ),
\label{loss}
\end{align}
where $y_k$ is the ground-truth label for the $k$-th action class.
\begin{figure}[t]
    \centering
    \includegraphics[width=0.5\textwidth]{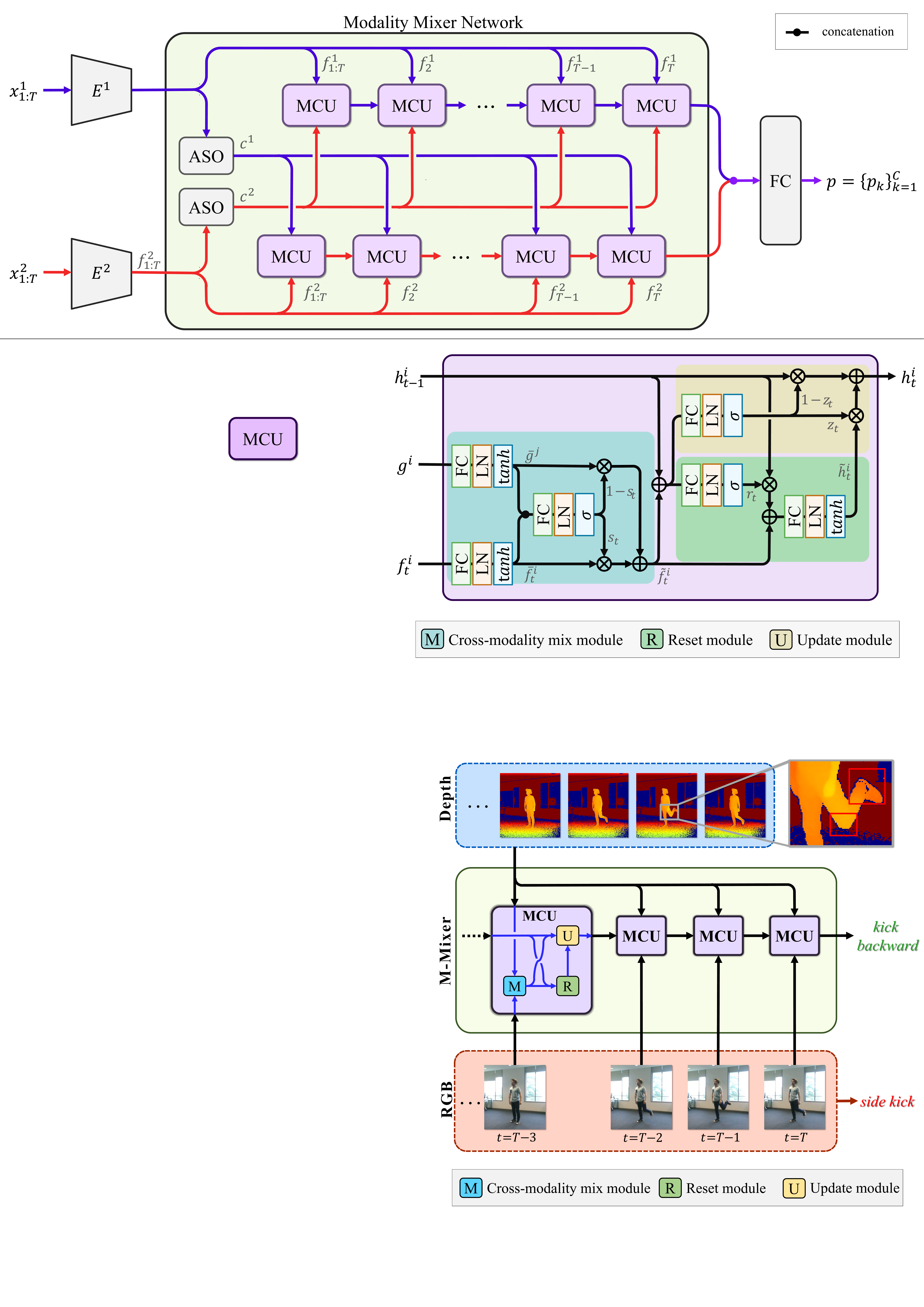}
    \caption{\textbf{Multi-modal Contextualization Unit (MCU).} 
    Our MCU consists of three modules: cross-modality mix module, reset module, and update module. 
    In a cross-modality mix module, a cross-action content $g^i$ is adaptively integrated with $f^i_t$, providing complementary information and the overall action content.
    Our reset module decides a reset gate $r_t$ to effectively drop and take information of previous hidden state $h^i_{t-1}$ and an integrated feature $\tilde{f}^i_{t}$.
    In an update module, an update gate $z_t$ is computed to update previous hidden state $h^i_{t-1}$.
    }
    \label{fig:mcu}
\end{figure}
\subsection{Multi-modal Contextualization Unit}
We describe our new recurrent unit, MCU, which is the core component of the proposed M-Mixer network.
Our MCU consists of three submodules: cross-modality mix module, reset module, and update module.
At the $t$-th timestep, the proposed MCU takes $f^i_t$ and $g^i$ to contextualize a modality-specific feature with a cross-modality action content feature.
This strategy enables MCU to supplement with complementary information of other modalities in terms of global action content.
As a result, the proposed MCU exploits rich and well-contextualized features for action recognition.

\paragraph{Cross-modality Mix Module.} First, ${f}^i_t$ and ${g}^i$ are projected to the same embedding space, as follows:
\begin{align}
    \bar{f}^i_t & = \tanh \left ( \texttt{LN} \left ( \mathbf{W}_f f^i_t \right ) \right ), \\
    \bar{g}^i &= \tanh \left (\texttt{LN} \left ( \mathbf{W}_g g^i \right )\right ),
\end{align}
where $\mathbf{W}_g \in \mathbb{R}^{((N-1)\times d_c) \times d_h)}$ and $\mathbf{W}_f \in \mathbb{R}^{d_f \times d_h}$ are trainable matrices, and $\texttt{LN}$ and $\tanh$ denote the layer normalization and the hyperbolic tangent function, respectively.
Note that we exclude a bias term for simplicity.

Next, an integration score $s_t$ is computed to determine how much representations of target modality and other modalities are activated, as follows:
\begin{align}
    s_t &= \sigma \left (\texttt{LN} \left (\mathbf{W}_s [\bar{f}^i_t \parallel \bar{g}^j] \right ) \right ),
\end{align}
where $\sigma$ indicates the sigmoid function and $\mathbf{W}_s \in \mathbb{R}^{2d_h \times d_h}$ is a weight matrix.
Then, $\bar{f}^i_t$ and $\bar{g}^i$ are combined to the supplemented feature $\tilde{f}^i_t$, as follows:
\begin{align}
    \tilde{f}^i_t& = s_t \otimes \bar{f}^i_t + \left (1-s_t \right ) \otimes \bar{g}^j,
\end{align}
where $\otimes$ denotes the element-wise multiplication.

\paragraph{Reset and Update Module.}
Our reset and update modules learn relationships between the supplemented feature $\tilde{f}^i_t$ and previous hidden state $h^i_{t-1}$.
In a reset module, a reset gate $r_t$ effectively drops and takes information of $h^i_{t-1}$ and $\tilde{f}^i_t$.
And an update module measures an update gate $z_t$ to amend previous hidden state $h^i_{t-1}$ to current hidden state $h^i_{t}$.
We compute $r_t$ and $z_t$, as follows:
\begin{align}
    r_t &= \sigma \left ( \texttt{LN} \left ( \mathbf{W}_r \left (  \tilde{f}^i_t + h^i_{t-1} \right ) \right ) \right ),\\
    z_t &= \sigma \left (\texttt{LN} \left (\mathbf{W}_z \left (\tilde{f}_t^i + h^i_{t-1} \right ) \right ) \right ),
\end{align}
where $\mathbf{W}_r \in \mathbb{R}^{d_h \times d_h}$ and $\mathbf{W}_z \in \mathbb{R}^{d_h \times d_h}$ are learnable parameters.
Then, the hidden state $h^i_{t-1}$ is updated with $z_t$, as follows:
\begin{align}
    h^i_t &= z_t \otimes \tilde{h}^i_t +\left (1-z_t \right )\otimes h^i_{t-1},
\end{align}
where $\tilde{h}$ is defined as:
\begin{align}
    \tilde{h}^i_t = \tanh \left (\texttt{LN} \left (\mathbf{W}_h\left (r_t \otimes h^i_{t-1} + \tilde{f}^i_t \right ) \right ) \right ).
\end{align}
Here, $\mathbf{W}_{h} \in \mathbb{R}^{d_h \times d_h}$ is a trainble matrix.

\section{Experiments}
\label{chap:exp}

\subsection{Dataset}
\paragraph{NTU RGB+D 60.}NTU RGB+D 60~\cite{ntu60} is a large-scale human action recognition dataset, consisting of 56,880 videos.
It includes 40 subjects performing 60 action classes in 80 different viewpoints.
As suggested in~\cite{ntu60}, we follow the cross-subject evaluation protocol.
For this evaluation, this dataset is split into 40,320 samples for training and 16,560 samples for testing.

\paragraph{NTU RGB+D 120.} As an extended version of NTU RGB+D 60, NTU RGB+D 120~\cite{ntu120} is one of the large-scale multi-modal dataset for video action recognition.
It contains 114,480 video clips of 106 subjects performing 120 classes from 155 different viewpoints.
We follow the cross-subject evaluation protocol as proposed in~\cite{ntu120}.
For the cross-subject evaluation, the 106 subjects are divided into 53 subjects for training and the remaining 53 subjects for testing.

\paragraph{Northwestern-UCLA (NW-UCLA).}
NW-UCLA~\cite{nwucla} is composed of 1475 video clips with 10 subjects performing 10 actions.
Each scenario is captured by three Kinect cameras at the same time from three different viewpoints.
As suggested in~\cite{nwucla}, we follow the cross-view evaluation protocol, using two views for training and the other one for testing.

\subsection{Implementation Details}
For the feature extractor for each modality, we use ResNet-18~\cite{resnet} for NTU RGB+D 60 and NW-UCLA and ResNet-34 for NTU RGB+D 120, which are initialized with ImageNet~\cite{imagenet} pretrained weights.
We set the size of the hidden dimension in MCU, $d_h$, to 512.
The input of each modality is a video clip uniformly sampled with temporal stride 8.
For the training procedure, we adopt random cropping and resize each frame to $224 \times 224$. We also apply random horizontal flipping and random color jittering for RGB videos.
For depth and IR frames, we use the same method to~\cite{garcia_admd}, a jet colormap, to convert those to color images.

To train our M-Mixer network, we use 4 GPUs of RTX 3090. 
We use the Adam~\cite{adam} optimizer with the initial learning rate of $10^{-4}$.
A batch size per GPU is 8 on NTU RGB+D 60 and NTU RGB+D 120.
Due to the small number of training samples, we use a single GPU with batch size 8 on NW-UCLA.

\subsection{Ablation Study}
In this section, we conduct extensive experiments to show the effectiveness of the proposed M-Mixer network and MCU.
All experiments in this section are conducted on NTU RGB+D 60~\cite{ntu60} with RGB and depth modalities.

\subsubsection{Action Summarizing Operator (ASO)}
\begin{table}[t]
\renewcommand{\arraystretch}{1.1}
\begin{center}
\begin{tabular}{c|c}
\hline
ASO &  Accuracy(\%) \\ \hline \hline
GRU~\cite{gru} &  90.66\\
Max pooling & 90.65\\
Average pooling &  90.77\\ \hline
\end{tabular}
\end{center}
\vspace{0.05cm}
\caption{\textbf{Ablation study of Action Summarizing Operator (ASO).} We use three instantiations of ASO: GRU, max pooling, and average pooling.}
\label{tab:soa}
\end{table}
\label{sec:exp:ablation of aso}
In Table~\ref{tab:soa}, we test three instantiations of Action Summarizing Operator (ASO).
For the average and max pooling, we apply mean and max operation across the temporal axis, respectively.
Also, we employ a GRU~\cite{gru} as ASO, which encodes a sequence into a vector on the time axis.
Our M-Mixer network achieves 90.66\% with GRU, 90.65\% with max pooling, and 90.77\% with average pooling.
We observe that there are very small performance differences between the three operations.
Among them, we empirically find that average pooling works slightly better than others.
Therefore, we utilize the average pooling as ASO in all experiments of this paper.

\subsubsection{Comparison with RNNs}
\begin{table}[t]
\renewcommand{\arraystretch}{1.1}
\begin{center}
\begin{tabular}{c|c}
\hline
Method & Accuracy (\%) \\ \hline \hline
LSTM~\cite{lstm}&  84.58\\ 
GRU~\cite{gru} & 84.87\\ \cdashline{1-2}
MCU & \textbf{90.77}\\ \hline
\end{tabular}
\end{center}
\caption{\textbf{Comparions with LSTM, GRU, and MCU.}
The best score is marked in \textbf{bold}.}
\label{tab:rnn_ablation}
\end{table}

To solely see the effectiveness of MCU, we replace our MCU in M-Mixer with LSTM~\cite{lstm} or GRU~\cite{gru}.
For these experiments, we use one LSTM or GRU for each modality.
We input a feature sequence $f^i_{1:T}$ to each LSTM or GRU and do not use the cross-modality action content.
The final predictions are calculated with concatenated output features of each LSTM or GRU, as same as our M-Mixer.
Table~\ref{tab:rnn_ablation} presents the performances of three networks.
We analyze the effect of the action content feature by comparing the results of LSTM, GRU, and MCU.
Compared to LSTM and GRU, our MCU learns relations between a current feature and the cross-modality action content to explore discriminative action information.
Thanks to the global video content and complementary information, the proposed MCU achieves performance gains of 6.19\% and 5.90\% over LSTM and GRU, respectively.

\subsubsection{Modality Contextualization Unit (MCU)}

\begin{table}[t]
\renewcommand{\arraystretch}{1.1}
\begin{center}
\begin{tabular}{c|ccc|c}
\hline
Exp. & \begin{tabular}[c]{@{}c@{}}Cross-modality\\ mix module\end{tabular} & \begin{tabular}[c]{@{}c@{}}Cross-modality\\ action content \end{tabular} & LN & \begin{tabular}[c]{@{}c@{}}Acc.\\ (\%)\end{tabular} \\ \hline \hline
\uppercase\expandafter{\romannumeral1}& \checkmark &  &  & 86.82\\
\uppercase\expandafter{\romannumeral2}& &  \checkmark&  &  87.76\\
\uppercase\expandafter{\romannumeral3}& \checkmark&  \checkmark&  &  89.97 \\
\uppercase\expandafter{\romannumeral4}& \checkmark&  \checkmark&  \checkmark& \textbf{90.77} \\ \hline
\end{tabular}
\end{center}
\caption{\textbf{Ablation study of MCU.}
MCU has three important components: a cross-modality mix module, cross-modality action content, and layer normalization indicated LN.
The best scores are marked in \textbf{bold}.}
\label{tab:mcu_ablation}
\end{table}

In this section, we validate the effects of three important components of our MCU: a cross-modality mix-module, the cross-modality action content, and the layer normalization.
To observe the abilities of each model component, we conduct ablation experiments on these three components and report the performances in Table~\ref{tab:mcu_ablation}.
In experiment \uppercase\expandafter{\romannumeral1}, we replace the cross-modality action content $g^i$ to the self-modality action content $c^i$ and turn off the layer normalization.
Experiment \uppercase\expandafter{\romannumeral2} is conducted to investigate the effect of the cross-modality mix module, where we change it to simple concatenation and disable the layer normalization.
Lastly, in experiment \uppercase\expandafter{\romannumeral3}, we only turn off the layer normalization of our MCU.
Comparing the result of experiment \uppercase\expandafter{\romannumeral1}, utilizing cross-modality action content achieves 89.97\% in experiment \uppercase\expandafter{\romannumeral3}, which is 3.15\%p higher than experiment \uppercase\expandafter{\romannumeral1}.
By comparing experiment \uppercase\expandafter{\romannumeral2} and \uppercase\expandafter{\romannumeral3}, we observe that using cross-modality mix module improves the performance from 87.76\% to 89.97\%.
Finally, we obtain the best performance of 90.77\% with all three components in experiment \uppercase\expandafter{\romannumeral4}.


\begin{table}[t]
\renewcommand{\arraystretch}{1.1}
\begin{center}
\begin{tabular}{c|cc|c}
\hline
\multirow{2}{*}{Modality} & \multicolumn{2}{c|}{Accuracy(\%)} & \multirow{2}{*}{$\Delta$(\%p)} \\ \cline{2-3}
 & MCU-self & MCU &  \\ \hline \hline
RGB &  56.61& \textbf{79.42} & + 22.81  \\
Depth &  84.31& \textbf{88.59} & + 4.25\\ \cdashline{1-4}
RGB+Depth &  88.17& \textbf{90.77} & + 2.60 \\ \hline
\end{tabular}
\end{center}
\caption{\textbf{Experiments on the effectiveness of the cross-modality action content.} For MCU-self, we use $c^i$ instead of $g^i$ to MCU.
A single modality represents the performance of each modality stream.
$\Delta$ indicates performance differences between MCU-self and MCU.
The best scores are marked in \textbf{bold}.}
\label{tab:cross-modal}
\end{table}
\begin{figure*}[t]
    \centering
    \includegraphics[width=\textwidth]{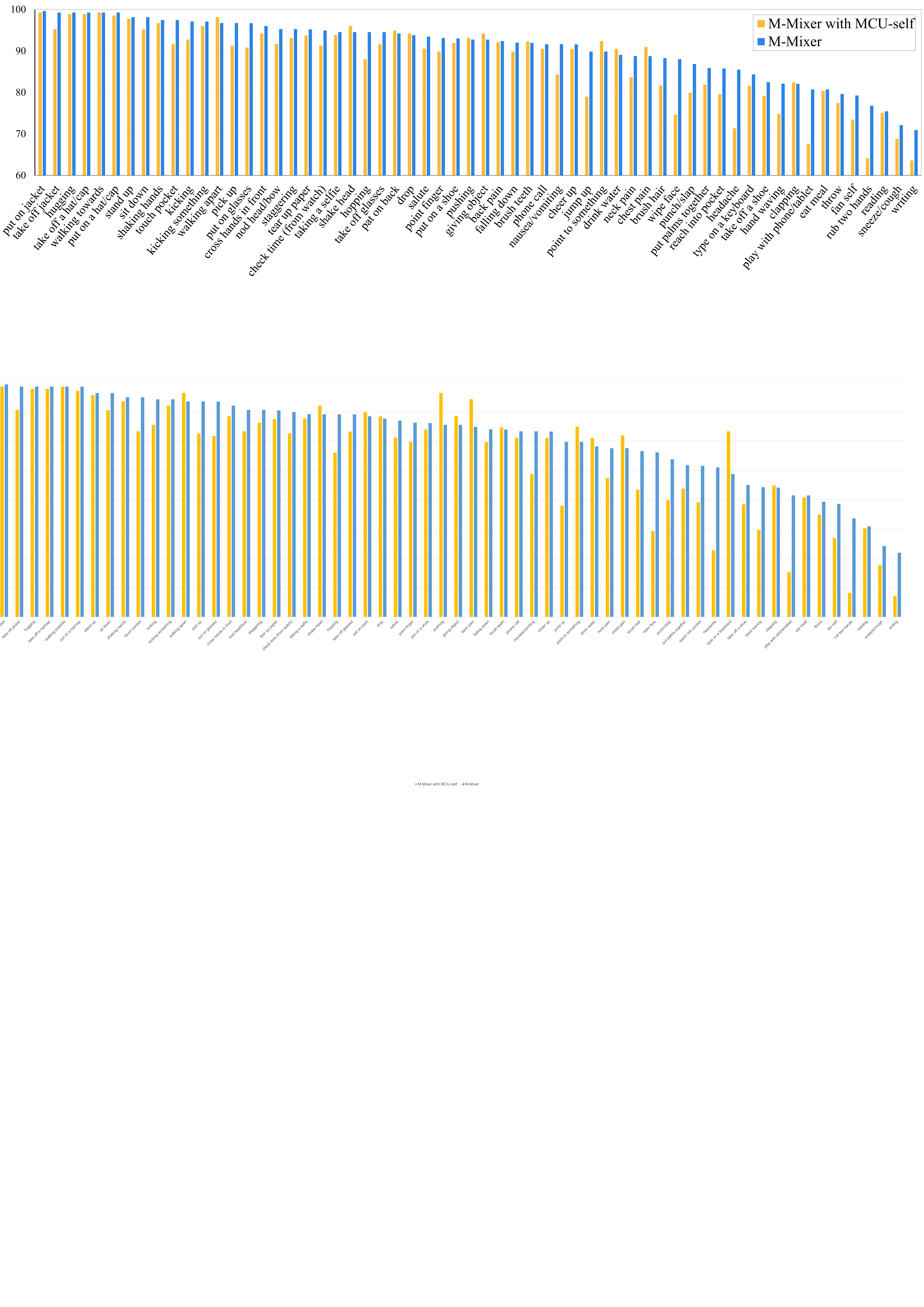}
    \caption{\textbf{Class-wise performance of M-Mixer network on NTU RGB+D 60~\cite{ntu60}.}
    60 action classes are listed in descending order according to the performance of our M-Mixer.
    }
    \label{fig:classwise}
\end{figure*}

\subsubsection{Cross-modality Action Content in MCU}
To validate the effectiveness of the cross-modality action content, we strategically replace the cross-modality action content of MCU (see Eq.~\ref{eq:gi}) to the self-modality action content (\ie, $c^i$).
Since the self-modality action content $c^i$ is from the same modality as a sequence $f^i_{1:T}$ to be encoded, the self-modality action content comprises global action information.
On the other hand, the cross-modality action content $g^i$ includes not only global action information but also complementary information of other modalities.
We name MCU with the self-modality action content MCU-self.

Furthermore, in order to closely analyze the effect of the cross-modality action content, we evaluate action recognition performances of a single modality in our M-Mixer with MCU and MCU-self.
In other words, we test the performance of RGB and depth features for employing MCU and MCU-self.
To this end, we train two additional fully-connected layers to classify an action class with $h^i_T$, as follows:
\begin{align}
    p^i = \textrm{softmax}\left (\mathbf{W}_{p^i} h^i_T\right ),
\end{align}
where $p^i$ is a probability distribution of $i$-th modality, and $\mathbf{W}_{p^i} \in \mathbb{R}^{d_h \times K}$ is a learnable matrix.
To train two classifiers, we use a loss function $L_h$ with the standard cross-entropy loss, as follows:
\begin{align}
    L_h = \sum^2_{i=1}\sum^K_{k=1} y_k \log \left ( p_k^i \right ),
\end{align}
where $p_k^i$ is a probability of $k$-th action class for $i$-th modality.
Note that the whole weights of M-Mixer network are fixed during training of the classifiers.

\paragraph{Comparative analysis in respect of modalities.}Table~\ref{tab:cross-modal} presents the results of comparative experiments about MCU and MCU-self.
With RGB and depth modalities, MCU-self obtains 88.17\%.
Meanwhile, our MCU achieves 90.77\%, which is 2.6\%p higher than the performance of MCU-self.
These results demonstrate the effectiveness of the cross-modality action content.
 
Compared to the self-modality action content, the cross-modality action content contains complementary information of other modalities as well as global action content.
Specifically, the RGB feature is strengthened with depth information, and the depth feature is augmented by RGB information in the setting of this experiment.
As a result, our MCU achieves 79.42\% in RGB stream and 88.59\% in depth stream, which are 22.81\%p and 4.28\%p higher than RGB and depth streams of MCU-self, respectively.
From these results, we demonstrate that the cross-modality action content effectively provides additional information across modalities and our MCU successfully utilizes complementary information in temporal encoding.

\paragraph{Comparison of class-wise performance.}In Fig.~\ref{fig:classwise}, we report class-wise performance of the proposed M-Mixer and M-Mixer with MCU-self.
60 action classes on NTU RGB+D 60~\cite{ntu60} are depicted in descending order based on the performances of our M-Mixer.
In most of the action classes, our M-Mixer achieves higher performances than using MCU-self.
Especially, the proposed M-Mixer has significant performance improvements in the action classes, which is lower in using MCU-self (\eg, `rub two hands', `headache', and `writing').

\subsection{Comparisons with state-of-the-arts}
\begin{table}[t]
\renewcommand{\arraystretch}{1.1}
\begin{center}
\begin{tabular}{c|c|c}
\hline
Method & Modality & Accuracy(\%) \\ \hline \hline
Sharoudy \etal~\cite{shahroudy2017deep} & RGB + Depth &74.86 \\
Liu \etal~\cite{liu2018viewpoint} & RGB + Depth &77.5 \\
ADMD~\cite{garcia_admd} & RGB + Depth & 77.74 \\
Dhiman \etal~\cite{mstd} & RGB + Depth & 79.4 \\
Garcia \etal~\cite{garcia_eccv} & RGB + Depth & 79.73 \\
c-ConvNet~\cite{rgbd_aaai} & RGB + Depth &86.42 \\
DMCL~\cite{garcia_dmcl}  & RGB + Depth + F & 87.25 \\
Wang \etal~\cite{mmar_lstm}  & RGB + Depth + F& 89.51 \\ \cdashline{1-3}
Ours (R18) & RGB + Depth &  \textbf{90.77}\\
Ours (R18) & RGB + Depth + IR &  \textbf{91.13}\\ \hline
\end{tabular}
\end{center}
\vspace{0.05cm}
\caption{\textbf{Performance Comparison on NTU RGB+D 60~\cite{ntu60}}. `F' denotes the optical flow. `R18' indicates our M-Mixer with ResNet18~\cite{resnet} feature extractor.
The best scores are marked in \textbf{bold}.}
\label{tab:ntu-60}
\end{table}

We compare our M-Mixer network with state-of-the-art methods on NTU RGB+D 60~\cite{ntu60}, NTU RGB+D 120~\cite{ntu120}, and NW-UCLA~\cite{nwucla} for multi-modal action recognition.

\vspace{-1.5mm}
\paragraph{NTU RGB+D 60.} In Table~\ref{tab:ntu-60}, we report performances of our M-Mixer and state-of-the-art approaches.
With RGB and depth modality, our M-Mixer achieves the best performance of 90.77\%, which is 1.18\%p and 3.44\%p higher than DMCL~\cite{garcia_dmcl} and the method proposed by Wang \etal~\cite{mmar_lstm}, respectively.
Note that those two methods use additional information of optical flow.
With RGB, depth, and IR videos, our M-Mixer obtains 91.13\%.
This result shows that the proposed M-Mixer can be extended to more than two modalities.

\vspace{-1.5mm}
\paragraph{NTU RGB+D 120.} Table~\ref{tab:ntu-120} shows performance comparisons on NTU RGB+D 120.
While NTU RGB+D 120 contains twice as many samples and classes as NTU RGB+D 60, our M-Mixer still obtains the state-of-the-art performance of 90.12\%.
Compared to the VGG architecture proposed in~\cite{ntu120}, our M-Mixer attains 28.2\% higher performance.
Also, M-Mixer achieves 0.38\% higher performance than DMCL.
\begin{figure*}[t]
    \centering
    \includegraphics[width=\textwidth]{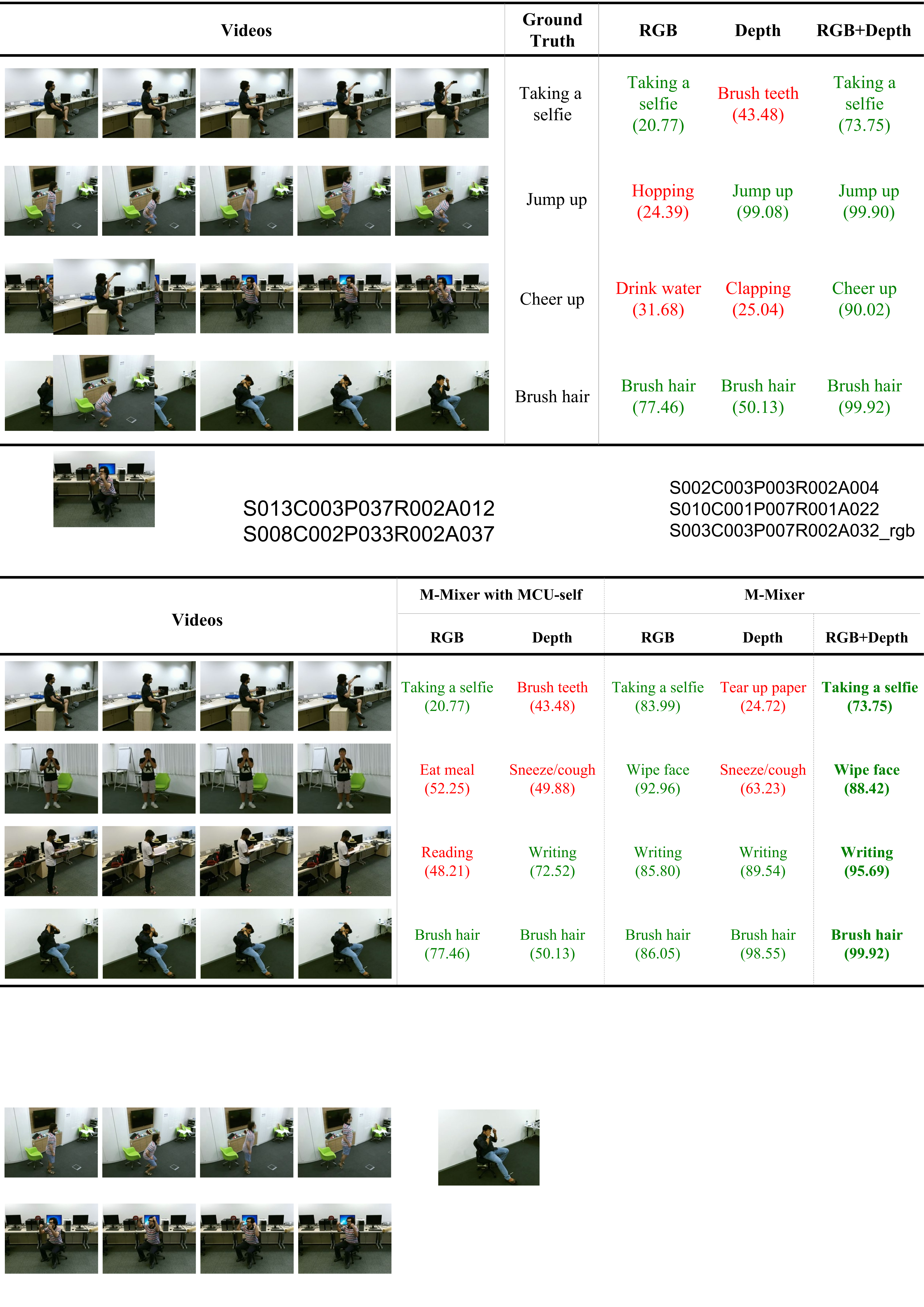}
    \caption{\textbf{Qualitative evaluation of M-Mixer network on NTU RGB+D 60~\cite{ntu60}.} 
    Predicted results consistent with ground-truth are colored in \textcolor{Green}{green}, otherwise in \textcolor{red}{red}. RGB, Depth, and RGB+Depth indicate prediction results form its respective stream.
    Also, confidence scores of predictions are presented in parentheses.}
    \label{fig:result}
\end{figure*}
\begin{table}[t]
\renewcommand{\arraystretch}{1.1}
\begin{center}
\begin{tabular}{c|c|c}
\hline
Method & Modality & Accuracy(\%) \\ \hline \hline
VGG~\cite{ntu120} & RGB+Depth & 61.9 \\
DMCL~\cite{garcia_dmcl} & RGB + Depth + Flow & 89.74 \\ \cdashline{1-3}
Ours (R34) & RGB+Depth &  \textbf{90.12}\\ \hline
\end{tabular}
\end{center}
\vspace{0.05cm}
\caption{\textbf{Performance Comparison on NTU RGB+D 120~\cite{ntu120}.} 
`R34' indicates our M-Mixer with ResNet34~\cite{resnet} feature extractor.
The best scores are marked in \textbf{bold}.}
\label{tab:ntu-120}
\end{table}
\begin{table}[t]
\renewcommand{\arraystretch}{1.1}
\begin{center}
\begin{tabular}{c|c|c}
\hline
Method & Modality & Accuracy(\%) \\ \hline \hline
Garcia \etal~\cite{garcia_eccv} & RGB + Depth & 88.87 \\
ADMD~\cite{garcia_admd} & RGB + Depth & 89.93 \\
Dhiman \etal~\cite{mstd} & RGB+Depth & 84.58 \\
DMCL~\cite{garcia_dmcl} & RGB + Depth + Flow &  93.79\\ \cdashline{1-3}
Ours (R18) & RGB+Depth &  \textbf{94.43}\\ \hline
\end{tabular}
\end{center}
\vspace{0.05cm}
\caption{\textbf{Performance Comparison on NW-UCLA~\cite{nwucla}.} 
`R18' indicates our M-Mixer with ResNet18~\cite{resnet} feature extractor.
The best scores are marked in \textbf{bold}.
}
\label{tab:nwucla}
\end{table}

\paragraph{NW-UCLA.}In Table~\ref{tab:nwucla}, we summarize the results on NW-UCLA.
Our M-Mixer network surpasses the state-of-the-art methods by achieving 94.43\%.
This performance is 0.64\% higher than DMCL that utilizes three modalities, and 9.85\% higher than the method proposed by Dhiman \etal~\cite{mstd} with RGB and depth.
This demonstrates the superiority of our M-Mixer network in discriminating actions.

\subsection{Qualitative Evaluation}
Figure~\ref{fig:result} shows the prediction results of M-Mixer on the sample videos of the NTU RGB+D 60~\cite{ntu60}.
To clearly see the efficacy of cross-modality action content, we also report the prediction results of M-Mixer with MCU-self.
In addition, we present the confidence score of each prediction under the predicted label.
We observe that our M-Mixer significantly improves the prediction results of both RGB and depth streams in comparison to using MCU-self.
For example, in the second row of~\ref{fig:result}, while RGB and depth of M-Mixer with MCU-self incorrectly predicts `wipe face' to `eat meal' and `sneeze/cough', RGB and RGB+Depth of M-Mixer classify the video correctly to ‘wipe face’.
Also, M-Mixer achieves higher confidence scores than M-Mixer with MCU-self for correctly predicted action classes (see the last row of~\ref{fig:result}).
These results show that using cross-modality action content is more effective in leveraging complementary information from other modalities than self-modality action content.
With these high performances of a single modality, our M-Mixer with all modalities successfully correctly predicts the action classes with high confidence scores.
From these results, we demonstrate the superiority of our proposed method.
More qualitative results are in our supplementary material.


\section{Conclusion}
\label{chap:conclusion}
In this paper, we have proposed the Modality Mixer (M-Mixer) network for multi-modal action recognition.
Also, we have introduced a novel recurrent unit, called Multi-modal Contextualization Unit (MCU), which is a key component of our M-Mixer network.
Our MCU deals with two important factors for multi-modal action recognition: 1) complementary information across modalities and 2) temporal context of the action.
Taking a feature sequence and a cross-modality content, MCU effectively learns the complementary relationships between modalities as well as the interactions between an action of the current timestep and the global action content.
In comprehensive ablation studies, we demonstrate the effectiveness of our proposed methods.
Moreover, we confirmed that our M-Mixer network outperforms state-of-the-art methods on NTU RGB+D 60~\cite{ntu60}, NTU RGB+D 120~\cite{ntu120}, and NW-UCLA~\cite{nwucla} for multi-modal action recognition.

\section*{Acknowledgment}
This work was supported by the Agency For Defense Development by the Korean Government (UD190031RD).

\clearpage
\newpage

{\small
\bibliographystyle{ieee_fullname}
\bibliography{egbib}
}

\end{document}